\documentclass[conference]{IEEEtran}
\IEEEoverridecommandlockouts
\usepackage{cite}
\usepackage{amsmath,amssymb,amsfonts}
\usepackage{algorithmic}
\usepackage{graphicx}
\usepackage{mathtools}
\usepackage{textcomp}
\usepackage{xcolor}
\usepackage{siunitx}
\usepackage{float}
\usepackage{pgfplots}
\DeclareUnicodeCharacter{2212}{−}
\usepgfplotslibrary{groupplots,dateplot}
\usetikzlibrary{patterns,shapes.arrows}
\pgfplotsset{compat=newest}
\usepackage{standalone}
\usepackage{cite}
\usepackage{amsmath,amssymb,amsfonts}
\usepackage{algorithmic}
\usepackage{amsthm}
\usepackage{svg}
\usepackage{makecell}
\usepackage{booktabs}
\usepackage{dblfloatfix}

\DeclareMathAlphabet{\mathcal}{OMS}{cmsy}{m}{n}

\DeclareMathOperator{\diag}{diag}
\DeclareMathOperator*{\argmin}{arg\,min}

\newcommand{\vect}[1]{\boldsymbol{\mathrm{#1}}}
\newcommand{\mat}[1]{\boldsymbol{\mathrm{#1}}}

\usepackage[acronym,shortcuts]{glossaries}
\makeglossaries
\newacronym[plural=BSs,firstplural=base stations (BSs)]{bs}{BS}{base station}
\newacronym[plural=PAs,firstplural=power amplifiers (PAs)]{pa}{PA}{power amplifier}
\newacronym{sdr}{SDR}{signal-to-distortion ratio}
\newacronym{snr}{SNR}{signal-to-noise ratio}
\newacronym{sndr}{SNDR}{signal-to-noise-and-distortion ratio}
\newacronym{snidr}{SNIDR}{signal-to-noise-and-interference-and-distortion ratio}
\newacronym{los}{LOS}{line-of-sight}
\newacronym{z3ro}{Z3RO}{zero third-order distortion}
\newacronym{mimo}{MIMO}{multiple input multiple output}
\newacronym{mrt}{MRT}{maximum ratio transmission}
\newacronym{dpd}{DPD}{digital pre-distortion}

\newacronym{ula}{ULA}{uniform linear array}

\newacronym{psd}{PSD}{power spectral density}
\newacronym{aclr}{ACLR}{adjacent channel leakage ratio}
\newacronym{cpu}{CPU}{central processing unit}
\newacronym{iid}{i.i.d.}{independently and identically distributed}
\newacronym{ue}{UE}{user equipment}
\newacronym{nlos}{NLoS}{non-line-of-sight}
\newacronym{hpbm}{HPBM}{half power beam width}
\newacronym{rts}{RTS}{ray tracing simulator}
\newacronym{ag}{AG}{array gain}
\newacronym{arp}{ARP}{Antenna Reference Point}
\newacronym{ecdf}{eCDF}{empirical cumulative distribution function}
\newacronym{evm}{EVM}{Error Vector Magnitude}
\newacronym{fft}{FFT}{fast Fourier transform}
\newacronym{ib}{IB}{in-band}
\newacronym{im}{IM}{intermodulation}
\newacronym{mf}{MF}{matched filtering}
\newacronym{mr}{MR}{maximum ratio}
\newacronym{ofdm}{OFDM}{orthogonal frequency division multiplexing}
\newacronym{oob}{OOB}{out-of-band}
\newacronym{papr}{PAPR}{peak-to-average power ratio}
\newacronym{rx}{RX}{Receiver}
\newacronym{trp}{TRP}{Transmission Reception Point}
\newacronym{tx}{TX}{transmitter}
\newacronym{zf}{ZF}{zero forcing}
\newacronym{dl}{DL}{downlink}
\newacronym{rzf}{RZF}{regularized zero forcing}
\newacronym{slc}{SLC}{spatial leakage suppression}
\newacronym{kpi}{KPI}{key performance indicator}
\newacronym{awgn}{AWGN}{additive white Gaussian noise}
\newacronym{qam}{QAM}{quadrature amplitude modulation}
\newacronym{amam}{AM/AM}{amplitude modulation to amplitude modulation}
\newacronym{ampm}{AM/PM}{amplitude modulation to phase modulation}
\newacronym{zmcscg}{ZMCSCG}{zero mean circularly symmetric complex Gaussian}

\newacronym{gnn}{GNN}{graph neural network}
\newacronym{ccnn}{CCNN}{circular convolutional neural network}
\newacronym{lrelu}{LReLU}{leaky rectified linear unit }
\newacronym{sdg}{SDG}{Sustainable Development Goal}
\newacronym{ibo}{IBO}{input back-off}
\newacronym{cdf}{CDF}{cumulative distribution function}
\newacronym{nn}{NN}{neural network}

\def\BibTeX{{\rm B\kern-.05em{\sc i\kern-.025em b}\kern-.08em
    T\kern-.1667em\lower.7ex\hbox{E}\kern-.125emX}}
\begin{document}

\title{Self-Supervised Learning of Linear Precoders under Non-Linear PA Distortion for Energy-Efficient Massive MIMO Systems\\
\thanks{This work was made possible by a mobility grant provided by the (Fonds wetenschappelijk Onderzoek) FWO with filenumber V424222N and a short-term scientific mission grant by the COST ACTION CA20120 INTERACT.}
}

\author{\IEEEauthorblockN{Thomas Feys*, Xavier Mestre\textsuperscript{\textdagger}, Fran\c{c}ois Rottenberg*
	}
	\IEEEauthorblockA{*KU Leuven, ESAT-WaveCore, Dramco, 9000 Ghent, Belgium
	}
    \IEEEauthorblockA{\textsuperscript{\textdagger}ISPIC, Centre Tecnologic Telecomunicacions Catalunya, Barcelona, Spain.}
}

\maketitle


\begin{abstract}
Massive \gls{mimo} systems are typically designed under the assumption of linear \glspl{pa}. However, \glspl{pa} are typically most energy-efficient when operating close to their saturation point, where they cause non-linear distortion. Moreover, when using conventional precoders, this distortion coherently combines at the user locations, limiting performance.  As such, when designing an energy-efficient massive \gls{mimo} system, this distortion has to be managed. In this work, we propose the use of a \gls{nn} to learn the mapping between the channel matrix and the precoding matrix, which maximizes the sum rate in the presence of this non-linear distortion. This is done for a third-order polynomial \gls{pa} model for both the single and multi-user case. By learning this mapping a significant increase in energy efficiency is achieved as compared to conventional precoders and even as compared to perfect \gls{dpd}, in the saturation regime.
\end{abstract}


\section{Introduction}
\subsection{Problem Formulation}
The estimated carbon footprint and electricity usage of the wireless communications sector continues to rise~\cite{trends2040}. As such, in the race to reduce carbon emissions and energy consumption by 2030 as stated by Europe's Green Deal~\cite{greendeal} and the United Nations \glspl{sdg}~\cite{sdgs}, the wireless communications sector is falling behind. In wireless communication systems, the \gls{pa} accounts for a large part of the energy consmption of a \gls{bs}~\cite{energy}. As such, it is vital to operate it in an energy-efficient manner. However, \glspl{pa} are most efficient close to their saturation point, where non-linear distortion arises. This leads to a trade-off between energy efficiency and linearity. In the past, linearity has taken the upper hand in this trade-off given that non-linear distortion limits the system capacity. As such, the \gls{pa} is typically operated at a certain back-off power in order to stay in the linear regime, which is detrimental for its energy efficiency. For the current technology, the energy efficiency of the \gls{pa} is typically as low as $5-30\%$~\cite{energy, pa_percentage}. In this work, we study how to operate the \glspl{pa} of a massive \gls{mimo} system closer to saturation by learning a precoding matrix that boosts performance in the presence of non-linear distortion. By doing so, the same capacity can be achieved while using less back-off, which improves the energy efficiency.

\subsection{State-of-the-Art}
As stated in the previous section, a large back-off is typically required to stay in the linear regime of the \gls{pa} which limits its energy efficiency. However, given the need to reduce energy consumption, this solution is no longer viable. Efforts to linearize the \gls{pa} such as \gls{dpd} are used in practical systems~\cite{pa_for_wireless }. However, \gls{dpd} techniques have a significant complexity burden, especially in massive \gls{mimo} where they have to be deployed at each antenna. Moreover, their performance is limited by clipping, i.e., the \gls{pa} can only be linearized up to the saturation point, so that a relatively large back-off is still required. More recent solutions incorporate knowledge of the distortion into the precoder design\cite{z3ro, zerofamily, z3ro_val, distortion-aware}. This allows for the spatial suppression of the distortion in the user directions, producing considerable gains over classical precoders. Unfortunately, these solutions are still limited in their practical implementation. In~\cite{distortion-aware} the solution to the precoding problem is obtained by solving a non-convex optimization problem with a projected gradient descent-based procedure. Given that the problem is non-convex, the procedure is executed multiple times in order to obtain a close-to-optimal solution. As an alternative solution for the problem, the authors in~\cite{z3ro} derived a, globally optimal, closed-form solution for the simplified single-user case and a \gls{los} channel, which was later extended to a general channel in~\cite{zerofamily}. There is thus a need for a solution that has low complexity and can address the challenging case of spatial user multiplexing.

\subsection{Contributions}

In this work, we propose the use of a \gls{nn} to find a mapping from the channel matrix to the precoding matrix. This mapping is learned under the presence of a non-linear \gls{pa} operated close to its saturation point, which introduces non-linear distortion. The need for machine learning arises from the non-linear and non-convex nature of the problem, which limits classical linear signal processing solutions. By learning the non-linear mapping from channel matrix to precoding matrix, a lot of the complexity is offloaded to the training step, which reduces the online computational complexity. This allows for a practical solution to the precoding problem under the presence of non-linear \gls{pa} distortion even in the multi-user case, which has not been addressed in previous works. As such, this opens perspectives to operate \glspl{pa} closer to their saturation point, which drastically increases their energy efficiency. 

\textbf{Notations}: Vectors and matrices are denoted by bold lowercase and bold uppercase letters respectively. Superscripts $(\cdot)^*$, $(\cdot)^{\intercal}$ and $(\cdot)^{H}$ stand for the conjugate, transpose and Hermitian transpose operators respectively. Subscripts $(\cdot)_m$ and $(\cdot)_k$ denote the antenna and user index. The expectation is denoted by $\mathbb{E}(.)$. The $M\times M$ identity matrix is given by $\mat{I}_M$. The main diagonal of a square matrix $\mat{A}$ is given by $\diag(\mat{A})$. The trace of a matrix is given by $\mathrm{Tr}\left(\cdot\right)$. The element-wise or Hadamard product of two matrices is denoted by $\mat{A} \odot \mat{B}$. The element at location $(i,j)$ in matrix $\mat{A}$ is indicated as $[\mat{A}]_{i,j}$.

\section{System Model}
In this work, we consider a massive \gls{mimo} system where the \gls{bs} is equipped with $M$ transmit antennas and $K$ single-antenna users are spatially multiplexed. The complex symbol intended for user $k$ is denoted as $s_k$ and is assumed to be zero mean circularly symmetric complex Gaussian with unit variance. The symbols between different users are assumed to be uncorrelated. The linearly precoded symbol at antenna $m$ is denoted by
\begin{align}
    x_m = \sum_{k=0}^{K-1} w_{m,k}s_k,
\end{align}
where $w_{m,k}$ is the precoding coefficient for user $k$ at antenna $m$. 
In matrix form, the precoded symbol vector $\vect{x} \in \mathbb{C}^{M\times1}$ is
\begin{align}
    \vect{x} = \mat{W}\vect{s},
\end{align}
where $\mat{W} \in \mathbb{C}^{M \times K}$ is the precoding matrix and $\vect{s} \in \mathbb{C}^{K \times 1}$ the symbol vector. The amplified transmit vector $\vect{y} \in \mathbb{C}^{M\times 1}$ is then given by 
\begin{align}
    \mat{y} = \phi\left(\vect{x}\right),
\end{align} 
where $\phi(\cdot)$ denotes the element-wise non-linear transformation caused by the \glspl{pa}.
The received signal vector $\vect{r} \in \mathbb{C}^{K\times1}$ is 
\begin{align*}
    \vect{r} &= \mat{H^\intercal} \vect{y} + \vect{v}
    =\mat{H^\intercal} \phi \left(\mat{W} \vect{s}\right) + \vect{v},
\end{align*}
with $\mat{H} \in \mathbb{C}^{M \times K}$ being the channel matrix. Each of its elements is assumed to be an \gls{iid} Rayleigh channel with zero mean and unit variance. The vector $\vect{v}\in \mathbb{C}^{K\times1}$ contains \gls{iid} zero mean complex Gaussian noise samples with variance $\sigma^2_v$.

\subsection{Modeling of PA Non-Linearities}\label{sec:pa}
In this work, the non-linear \gls{pa} is modeled as a third-order complex valued polynomial~\cite{pas}. The output of the \gls{pa} at antenna $m$ is given by
\begin{align}\label{eq:poly}
    \phi(x_m) = \beta_1 x_m + \beta_3 x_m |x_m|^2,
\end{align}
where $\beta_1$ and $\beta_3$ are complex coefficients that model both \gls{amam} and \gls{ampm} distortion. This third-order model is valid as the \gls{pa} enters saturation, given that the higher-order polynomial terms have a small contribution in this regime. 


\begin{figure*}[tb]
    \centering
    \includegraphics[width=\textwidth]{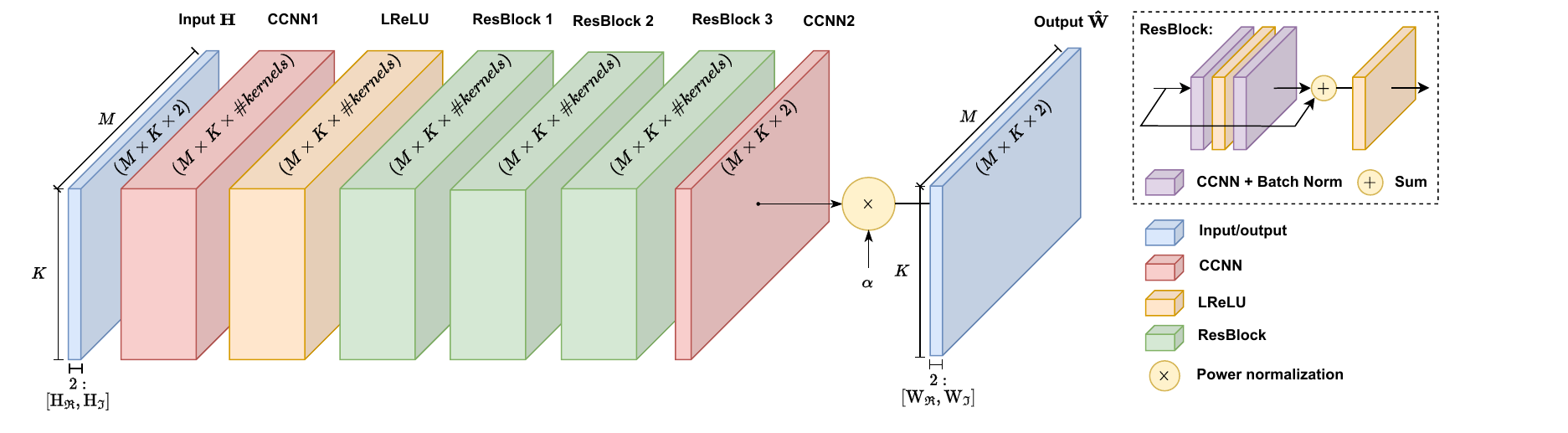}
    \caption{Circular convolutional neural network architecture.}
    \label{fig:ccnn}
\end{figure*}

\subsection{Optimization Problem}

An achievable sum rate $R_{\mathrm{sum}}$, i.e., a lower bound on the capacity, can be obtained by considering that the noise and distortion are jointly Gaussian distributed and independent from the data symbols, which can be seen as a worst case
\begin{align}\label{eq:rate}
        R_{\mathrm{sum}} = \sum_{k=0}^{K-1} \mathrm{log}_2(1 + \mathrm{SNIDR}_k),
\end{align}
where $\mathrm{SNIDR}_k$ is the \gls{snidr} at user $k$. It can be computed based on the Bussgang decomposition~\cite{demir2020bussgang}, which states that the received signal for user $k$ can be written as $  r_{k} = {B_{k}  s_k} + d_{k} + v_{k}$. Here, $d_k$ captures both the non-linear distortion and inter-user interference, which is uncorrelated to the transmit signal $s_k$ and the noise $v_k$. The linear gain is given by $B_{k}=\mathbb{E}\left(r_{k} s_k^{*}\right) / p_k$, with $p_k = \mathbb{E}(s_ks_k^*)$~\cite{demir2020bussgang}. The received signal variance for user $k$ is given by $|B_k|^2p_k$. The distortion and inter-user interference can be computed as $\mathbb{E}\left(|d_{k} |^{2}\right)=\mathbb{E}\left(|r_{k} |^{2}\right)-|B_{k} |^{2} p_k-\sigma_{v_{k}}^{2}$, given that $d_k$, $s_k$ and $v_k$ are uncorrelated. The \gls{snidr} for user $k$ is then given by
\begin{align}
    \mathrm{SNIDR}_k = \frac{|B_k |^{2} p_k}{\mathbb{E}\left(|d_{k}|^{2}\right)+\sigma_{v}^{2}}.
\end{align}
This general expression for the \gls{snidr} can be evaluated numerically and is used for the simulations in section~\ref{sec:sim}. 

For training the \gls{nn}, the specific case of a third-order \gls{pa} model is assumed, which simplifies the expression. By applying Bussgang's theorem\cite{demir2020bussgang} to the amplification stage, we can write the amplified signal as
\begin{align}
    \phi(\vect{x}) = \mat{G} \vect{x} + \vect{e},
\end{align}
with $\vect{e}\in \mathbb{C}^{M\times1}$ the non-linear distortion term and $\mat{G}\in \mathbb{C}^{M\times M}$ a diagonal matrix containing the Bussgang gains with the diagonal entries being $[\mat{G}]_{m,m} = \mathbb{E}[\phi(x_m)x_m^*]/\mathbb{E}[|x_m|^2]$. When assuming the third-order polynomial model given in equation~(\ref{eq:poly}) and linear precoding ($\vect{x} = \mat{W}\vect{s}$), we can write the gain matrix $\mat{G}$ as a function of the precoding matrix $\mat{W}$~\cite{distortion-aware}
\begin{align}
    \mat{G}(\mat{W}) &= \beta_1 \mat{I}_M + 2 \beta_3 \diag(\mat{C}_x).
\end{align}
This expression is valid when assuming that all PAs have the same polynomial coefficients. The input covariance matrix is given by $\mat{C_x} = \mathbb{E}[\vect{x}\vect{x}^H] = \mat{W}\mat{W}^H$. From~\cite{distortion-aware}, the covariance matrix of the non-linear distortion $\vect{e}$ can be derived as
\begin{align}
    \mat{C}_e(\mat{W}) = 2 |\beta_3|^2 \mat{C}_x \odot |\mat{C}_x|^2.
\end{align}
The received signal at user $k$ can then be written as
\begin{align}
    r_k &= \underbrace{\vect{h}_k^\intercal \mat{G(W)} \vect{w}_k s_k}_{\text{desired signal}} + \underbrace{\sum_{k'\neq k} \vect{h}_k^{\intercal} \mat{G(W)}\vect{w}_{k'}s_{k'}}_{\text{inter-user interfernce}} \\ 
    &+ \underbrace{\vect{h}_k^{\intercal} \vect{e}}_{\substack{\text{received} \\ \text{non-linear distortion}}} + \underbrace{v_k}_{\text{noise}}.
\end{align}
This leads to the following \gls{snidr} expression for user $k$
\begin{align}
    &\mathrm{SNIDR}_k(\mat{W}) = \frac{|\vect{h}_k^{\intercal} \mat{G(W)} \vect{w}_k|^2}{\sum\limits_{k'\neq k}|\vect{h}_k^{\intercal} \mat{G(W)} \vect{w}_{k'}|^2 + \vect{h}_k^{\intercal} \mat{C}_e(\mat{W}) \vect{h}^*_k + \sigma_v^2}.
\end{align}
Given this expression for the \gls{snidr}, an achievable sum rate can be computed using eq. (\ref{eq:rate}).
As such, the optimization problem we aim to solve can be formulated as
\begin{align}\label{eq:opt}
    \max_{\mat{W}} \quad & R_{\mathrm{sum}}\left(\mat{W}\right) \\
    \textrm{s.t.} \quad & \mathbb{E}\left(\sum_{m=0}^{M-1}|x_m|^2\right) =  \mathrm{Tr}\left(\mat{W}\mat{W}^H\right) \leq P_T, \nonumber
\end{align}
where $P_T$ is the total transmit power. The aim is thus to find a precoding matrix which maximizes the sum rate, subject to a power constraint\footnote{For simplicity, the power constraint is taken before the \gls{pa}, which neglects the non-linearly amplified power, which is small compared to the full transmit power.}, while the system is affected by PA non-linearities.


\section{Neural Network-Based Precoder}
Given the optimization problem defined in equation (\ref{eq:opt}), we propose a \gls{nn} $f: \mathbb{C}^{M \times K} \mapsto\mathbb{C}^{M \times K}$ which learns a mapping from channel matrix $\mat{H}$ to precoding matrix $\hat{\mat{W}}$. The \gls{nn} represents a learned non-linear function 
\begin{align}
    \hat{\mat{W}} = f\left(\mat{H}; \mat{\theta}\right),
\end{align}
where $\mat{\theta}$ are the learned parameters of the \gls{nn}.

\subsection{Neural Network Architecture}
A fully connected neural network can be used to learn the mapping between channel matrix $\mat{H}$ and precoding matrix $\hat{\mat{W}}$. However, when $M$ becomes large, the precoding problem becomes high dimensional. Additionally, the architecture of fully connected neural networks is structurally very general due to the high number of tunable parameters, which makes the training of these networks very complex. Hence, it is beneficial to select a neural network architecture which has a structure (i.e., inductive bias) that suits the learning task. The inductive bias of a network constrains the functions which can be learned, reducing the size of the hypothesis space covered by the \gls{nn}. When the inductive bias matches the learning task, i.e., the desired function can still be well approximated by the selected architecture, the learning performance can be improved while the training complexity is reduced~\cite{inductivebias}. 

From~\cite{cnn_gnn}, we know that the precoding task is permutation equivariant with respect to users and antennas, i.e., if the order of the users in the channel matrix $\mat{H}$ changes, the order of the precoding vectors changes accordingly but the sum rate stays the same. The same analysis is valid when the order of the antennas changes. \glspl{nn} that fit this property are the \gls{gnn} and \gls{ccnn} which were proposed for precoding in~\cite{cnn_gnn}.

In this work, the \gls{ccnn} illustrated in Figure~\ref{fig:ccnn} is used to learn the precoding task. This network consists of a \gls{ccnn} layer with \gls{lrelu} activation, followed by three residual blocks, a final \gls{ccnn} layer and a power normalization layer to satisfy the power constraint. Each residual block consists of a \gls{ccnn} layer followed by a batch normalization layer, a \gls{lrelu} activation and another \gls{ccnn} layer, after which this output is added to the input of the block and a final \gls{lrelu} activation is applied. The skip connections used in these residual blocks ensure stable gradients during training\cite{resnet}.  Each \gls{ccnn} layer, has a kernel size of $9\times 3$ and learns 256 kernels, except for the final \gls{ccnn} layer where only two kernels are learned in order to produce the desired output shape of $M \times K \times 2$ where the final dimension represents the real and imaginary parts of the precoding matrix. Note that the kernel size of the convolutions is selected to ensure global receptive field, meaning that each output feature (i.e., element of the precoding matrix) depends on the entire input (i.e., the full channel matrix)~\cite{receptivefield}. A kernel size of $9\times3$ produces a receptive field of $65\times17$ (computed using the open source library from~\cite{receptivefield}). The slope coefficient of the \gls{lrelu} activation for all layers is set to 0.01. The power normalization layer consists of a scalar normalization given by
 \begin{align}
    \mat{\hat{W}}^{norm} = \alpha \mat{\hat{W}},
\end{align}
with $\alpha = \sqrt{P_T/\mathrm{Tr}(\mat{\hat{W}} \mat{\hat{W}}^H)}$.

\subsection{Training and Hyperparameter Selection}
The \gls{nn} $f\left(\mat{H}; \mat{\theta}\right)$ is trained in a self-supervised manner by maximizing the sum rate in order to obtain the \gls{nn} parameters
\begin{align}
    \mat{\theta}^* = \argmin_{\mat{\theta}} - R_{\mathrm{sum}}(f\left(\mat{H}; \mat{\theta}\right)).
\end{align}
The parameters of the \gls{nn} are updated using the Adam optimizer\cite{adam}. The training set consists of 200000 generated Rayleigh fading channels sampled from a complex normal distribution with zero mean and variance one $[\mat{H}]_{i,j} \sim \mathcal{CN}(0,1)$. The hyperparameters are selected by using a validation set of size 2000, while for the simulations performed in section~\ref{sec:sim}, an independent test set of size 10000 is used. For training, a batch size of 256 is used, with an initial learning rate of $5\times10^{-3}$, which is reduced if the validation loss reaches a plateau. The network is trained for 50 epochs with early stopping if the validation loss does not further decrease.

\begin{table}[h]
\centering
    \caption[Caption for LOF]{\gls{pa} parameters at different back-off values ($\beta_1=1$).}
    \label{tab:paparams}
    \centering
\resizebox{\columnwidth}{!}{%
 \begin{tabular}{@{}cccccccc@{}} 
 \toprule
 IBO [\SI{}{\decibel}] & -9 & -7.5 & -6 & -4.5 & -3 & -1.5 & 0 \\ [0.5ex] 
     \midrule

 $\beta_3\,(\cdot10^{-3})$  &  \makecell{-19.93 \\ -10.80j} &  \makecell{-30.69 \\ -18.85j} &  \makecell{-42.26\\-24.91j} &  \makecell{-56.13\\-30.06j} &  \makecell{-77.82\\-40.12j} &  \makecell{-117.3\\-65.01j} &  \makecell{-159.1\\-79.25j}\\ 
  \bottomrule
\end{tabular}
}
\end{table}

\section{Simulations}\label{sec:sim}
For the following simulations, the polynomial coefficients, which are assumed to be equal across all \glspl{pa}, are obtained by a least squares regression of the third-order model to the modified Rapp model~\cite{modified_rapp}. The modified Rapp model contains both third and higher-order effects. The \gls{amam} and \gls{ampm} distortion of this model are 
\begin{align}
    \phi_{\mathrm{AM-AM}}(x_m) &= \frac{|x_m|}{\left(1+\left|\frac{x_m}{\sqrt{p_{\mathrm{sat}}}}\right|^{2S}\right)^{\frac{1}{2S}}}\\
    \phi_{\mathrm{AM-PM}}(x_m) &= \frac{A |x_m|^q}{1 + \left|\frac{x_m}{B}\right|^q}.
\end{align}
The \gls{amam} and \gls{ampm} distortion are then applied to the signal as follows
\begin{align}
    \phi(x_m) = \phi_{\mathrm{AM-AM}}(x_m)  e^{j(\angle x_m + \phi_{\mathrm{AM-PM}}(x_m))}.
\end{align}
In order to obtain the polynomial coefficients in Table~\ref{tab:paparams}, the modified Rapp model coefficients are set as follows\footnote{Adapted from~\cite{3gpp} to a \gls{pa} with unit gain.}: $S=2$, $q=4$, $A=-0.315$, $B=1.137$ and the saturation power of the \gls{pa} $p_{\mathrm{sat}}$ is scaled in order to produce the desired \gls{ibo} according to $\mathrm{IBO} = p_{\mathrm{in}}/p_{\mathrm{sat}}$, with $p_{\mathrm{in}}$ being the average input power at each \gls{pa}. 
Additionally, the proposed solution is compared against a perfect \gls{dpd}. This can be modeled as a linear \gls{amam} characteristic up to a certain saturation point where the output of the \gls{pa} is clipped~\cite{pas}. The \gls{amam} characteristic is thus modeled as
\begin{align}\label{eq:dpd}
     \phi_{\mathrm{AM-AM}}(x_m) &= \begin{cases}|x_m| & \text { for } \quad |x_m| \leq \sqrt{p_{\mathrm{sat}}} \\ \sqrt{p_{\mathrm{sat}}} & \text { for } \quad |x_m|>\sqrt{p_{\mathrm{sat}}}\end{cases},
\end{align}
while the \gls{ampm} conversion is zero.

Furthermore, for all simulations, the total transmit power is $P_T = M$. Hence, the average power at the input of each \gls{pa} is $p_{\mathrm{in}} = P_T/M = 1$. The linear \gls{pa} gain is set to one. Training and testing are done at an \gls{ibo} of \SI{-3}{\decibel}, unless specified otherwise, this saturates the \glspl{pa} much more than current cellular systems that require 9-12 \SI{}{\decibel} back-off. The polynomial coefficients corresponding to the \gls{ibo} value can be found in Table~\ref{tab:paparams}. During training, $P_T/\sigma^2_v$ is set to \SI{20}{\decibel}. After training, the \glspl{nn} are evaluated based on the sum rate given in (\ref{eq:rate}). 


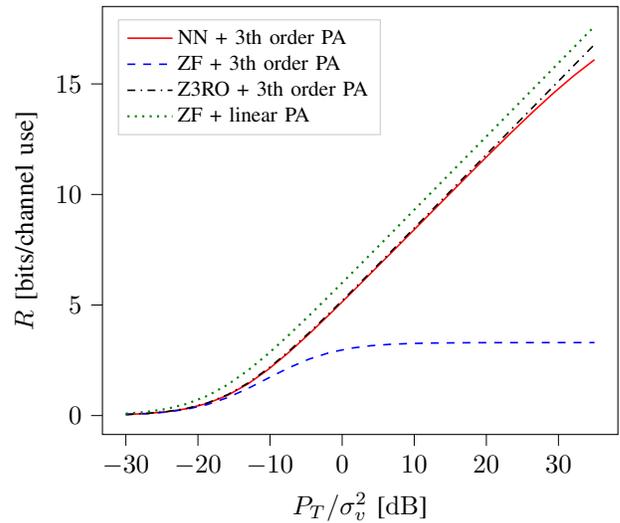
\begin{figure}[t]
    \centering
\begin{tikzpicture}

\definecolor{crimson2143940}{RGB}{214,39,40}
\definecolor{darkgray176}{RGB}{176,176,176}
\definecolor{darkorange25512714}{RGB}{255,127,14}
\definecolor{forestgreen4416044}{RGB}{44,160,44}
\definecolor{lightgray204}{RGB}{204,204,204}
\definecolor{steelblue31119180}{RGB}{31,119,180}
\definecolor{green01270}{RGB}{0,127,0}

\begin{axis}[
legend cell align={left},
legend style={nodes={scale=0.8, transform shape},
  fill opacity=0.8,
  draw opacity=1,
  text opacity=1,
  at={(0.03,0.97)},
  anchor=north west,
  draw=lightgray204
},
tick align=outside,
tick pos=left,
x grid style={darkgray176},
xlabel={$P_T / \sigma^2_v$ [\SI{}{\decibel}]},
xmin=-33.25, xmax=38.25,
xtick style={color=black},
y grid style={darkgray176},
ylabel={$R$ [bits/channel use]},
ymin=-0.854630579231094, ymax=18.4991359963405,
ytick style={color=black}
]
\addplot [semithick, red]
table {%
-30 0.0491030216217041
-27.1739139556885 0.0926780700683594
-24.3478260040283 0.172640204429626
-21.5217399597168 0.314376354217529
-18.6956520080566 0.552026152610779
-15.8695650100708 0.920033574104309
-13.043478012085 1.43689215183258
-10.2173910140991 2.09359097480774
-7.39130449295044 2.85943865776062
-4.5652174949646 3.69858050346375
-1.73913037776947 4.58192253112793
1.08695650100708 5.49006128311157
6.73913049697876 7.33928775787354
18.043478012085 11.0551662445068
20.8695659637451 11.9720392227173
23.6956520080566 12.8746652603149
26.5217399597168 13.7535533905029
29.3478260040283 14.5951204299927
32.1739120483398 15.3818683624268
35 16.0941200256348
};
\addlegendentry{NN + 3th order PA}
\addplot [semithick, blue, dashed]
table {%
-30 0.0457543134689331
-27.1739139556885 0.0861542224884033
-24.3478260040283 0.159788370132446
-21.5217399597168 0.288687825202942
-18.6956520080566 0.500025749206543
-15.8695650100708 0.81476092338562
-13.043478012085 1.22854852676392
-10.2173910140991 1.70020878314972
-7.39130449295044 2.1635947227478
-4.5652174949646 2.55708765983582
-1.73913037776947 2.8489146232605
1.08695650100708 3.04167079925537
3.91304349899292 3.15815901756287
6.73913049697876 3.22439312934875
9.56521701812744 3.26065063476562
12.3913040161133 3.28006601333618
18.043478012085 3.29573392868042
29.3478260040283 3.30121207237244
35 3.30153203010559
};
\addlegendentry{ZF + 3th order PA}

\addplot [semithick, black, dashdotted]
table {%
-30 0.0511907339096069
-27.1739139556885 0.0965514183044434
-24.3478260040283 0.179633736610413
-21.5217399597168 0.3264399766922
-18.6956520080566 0.57142972946167
-15.8695650100708 0.948466539382935
-13.043478012085 1.47445714473724
-10.2173910140991 2.13876914978027
-7.39130449295044 2.9100980758667
-4.5652174949646 3.75295805931091
-1.73913037776947 4.63898754119873
1.08695650100708 5.54945278167725
3.91304349899292 6.47324371337891
9.56521701812744 8.3388204574585
23.6956520080566 13.0284986495972
35 16.7835597991943
};
\addlegendentry{Z3RO + 3th order PA}

\addplot [thick, green01270, dotted]
table {%
-30 0.0888420343399048
-27.1739139556885 0.165701270103455
-24.3478260040283 0.302370548248291
-21.5217399597168 0.532628297805786
-18.6956520080566 0.891448020935059
-15.8695650100708 1.39892637729645
-13.043478012085 2.04782867431641
-10.2173910140991 2.80832672119141
-7.39130449295044 3.64444780349731
-4.5652174949646 4.5265965461731
-1.73913037776947 5.43492603302002
1.08695650100708 6.3575701713562
6.73913049697876 8.2222204208374
20.8695659637451 12.9115619659424
35 17.6054096221924
};
\addlegendentry{ZF + linear PA }
\end{axis}
\end{tikzpicture}
    \caption{Achievable rates averaged over 500 channel realizations taken from the test set, for $M=64$ and $K=1$. Comparing the \gls{nn} precoder against \gls{zf} (MRT) and the Z3RO precoder from~\cite{z3ro}.}
    \label{fig:singleuser}
\end{figure}

\begin{figure}[t]
    \centering
    \input{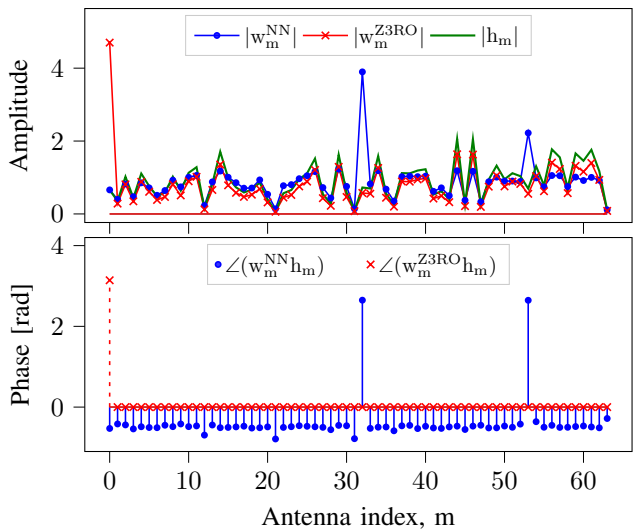}
    \caption{Modulus of channel and precoding coefficients for the \gls{nn} and Z3RO (top). Phase of the respective precoding coefficients multiplied with the channel coefficient (bottom). Both precoders saturate one (or a few) antennas with opposite phase shifts, to cancel distortion.}
    \label{fig:pwr}
\end{figure}

\subsection{Single-User Case}
In this section, the single-user scenario is considered ($K=1$), and the \gls{zf} precoder becomes the conventional \gls{mrt} precoder as no user interference needs to be canceled. The \gls{zf} precoder is designed under the assumption of a linear \gls{pa}. As a consequence, when a non-linear \gls{pa} is present, the user performance will be degraded as compared to the linear case. This can be seen in Fig.~\ref{fig:singleuser}, where the achievable rate of  the \gls{zf} precoder is depicted as a function of $P_T/\sigma^2_v$ when considering a linear \gls{pa} and the non-linear \gls{pa} model from~(\ref{eq:poly}). In the single-user case, the Z3RO precoder from~\cite{z3ro} provides a solution that mitigates the third-order distortion in the user direction, which comes at the cost of a small reduction in array gain. The Z3RO precoder saturates one (or a few) of the antennas with an opposite phase shift. When comparing the \gls{nn} precoder against the Z3RO precoder in Fig.~\ref{fig:singleuser}, we see that the \gls{nn} achieves similar performance as the Z3RO precoder, i.e., the rate is not limited by distortion but grows linearly with $P_T/\sigma^2_v$. Indeed, when comparing the amplitude and phase of both precoders in Fig.~\ref{fig:pwr}, it is clear that the \gls{nn} also saturates one or a few of the antennas with an opposite phase shift. In conclusion, for $K=1$, the \gls{nn} has learned a similar precoding structure as the Z3RO precoder. 


\subsection{Multi-User Case}
\begin{figure}[t]
    \centering
    \input{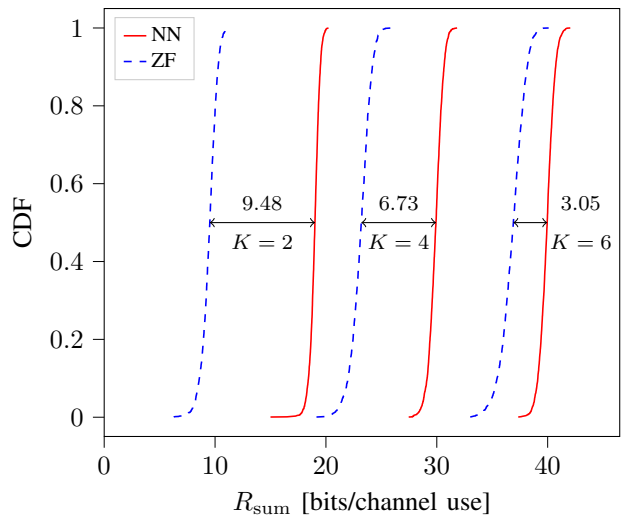}
    \caption{CDF of the sum rate over 2000 channel realizations taken from the test set, for $M=64$, $K \in \{2, 4, 6\}$ and $P_T/\sigma^2_v=20$ \SI{}{\decibel}. Comparing the \gls{nn} precoder against \gls{zf} for the third-order PA model.}
    \label{fig:cdf}
\end{figure}

When multiple users are present, both the inter-user interference and distortion to multiple users have to be mitigated. In this more complex scenario, there is no closed-form solution available for the optimal precoder. As such, the \gls{nn} is trained to learn how to perform this task. In Fig.~\ref{fig:cdf}, a comparison with the \gls{zf} precoder is made. In this figure, the \gls{cdf} of the sum rate is depicted for 2000 channel realizations, when using the polynomial \gls{pa} model from~(\ref{eq:poly}), for $K \in \{2,4,6\}$. 
When $K=2$ we see an increase in sum rate of 9.48 bit/channel use, when using the proposed method, 6.73 bit/channel use when $K=4$ and 3.05 bit/ channel use if $K=6$. This shows the ability of the \gls{nn} to cancel non-linear distortion in the multi-user scenario, which results in significant increases in capacity.  Additionally, this illustrates that the higher the number of users becomes, the less gain is to be obtained by using the \gls{nn} precoder. This is to be expected as non-linear distortion is more spatially spread out when more users are present~\cite{distortion_beamformed2}. In other words, less distortion is beamformed in the user directions, which leads to less potential gains for mitigating this distortion. Moreover, when more users are present, canceling all distortion to all users becomes more complex. Hence, the solution found by the \gls{nn} might not be the globally optimal one. Nevertheless, the \gls{nn}-based precoder achieves a significant increase in channel capacity as compared to the classical \gls{zf} precoding.

In Fig.~\ref{fig:multiuser}, the sum rate is depicted as a function of $P_T/\sigma^2_v$ for $K \in \{2, 4\}$. It is shown that the proposed solution can outperform a perfect \gls{dpd} (\ref{eq:dpd}) combined with \gls{zf}, when the system is highly distortion limited (i.e., for high \gls{snr} levels). This can be explained by the fact that a perfect \gls{dpd} can only account for weakly non-linear effects, i.e., the \gls{pa} can only be linearized up to the saturation point, after which clipping occurs. 

\begin{figure}[t]
    \centering
\begin{tikzpicture}

\definecolor{crimson2143940}{RGB}{214,39,40}
\definecolor{darkgray176}{RGB}{176,176,176}
\definecolor{darkorange25512714}{RGB}{255,127,14}
\definecolor{forestgreen4416044}{RGB}{44,160,44}
\definecolor{lightgray204}{RGB}{204,204,204}
\definecolor{steelblue31119180}{RGB}{31,119,180}

\begin{axis}[
legend cell align={left},
legend style={nodes={scale=0.8, transform shape},fill opacity=0.8, draw opacity=1, text opacity=1, draw=lightgray204},
tick align=outside,
tick pos=left,
x grid style={darkgray176},
xlabel={$\mathrm{IBO}=p_{\mathrm{in}}/p_{\mathrm{sat}}$ [dB]},
xmin=-9.45, xmax=0.45,
xtick style={color=black},
y grid style={darkgray176},
ylabel={$R_{\mathrm{sum}}$ [bits/channel use]},
ymin=0, ymax=45,
ytick style={color=black}
]
\addplot [semithick, red, mark=square, mark size=1, mark options={solid}]
table {%
-9 37.8049087524414
-7.5 35.1609764099121
-6 32.6603240966797
-4.5 30.2455310821533
-3 30.217586517334
-1.5 26.9535026550293
0 24.4091110229492
};
\addlegendentry{NN}
\addplot [semithick, blue, dashed, mark=*, mark size=1, mark options={solid}]
table {%
-9 37.1434860229492
-7.5 33.4332847595215
-6 30.1344528198242
-4.5 27.1424255371094
-3 23.1284561157227
-1.5 17.2662315368652
0 12.941460609436
};
\addlegendentry{ZF}
\addplot [semithick, black, dotted, mark=x, mark size=2, mark options={solid}]
table {%
-9 41.2644386291504
-7.5 38.8961868286133
-6 34.5880889892578
-4.5 30.1036243438721
-3 26.1978397369385
-1.5 23.0889072418213
0 20.7224235534668
};
\addlegendentry{ZF+DPD}

\addplot [semithick, red, mark=square, mark size=1, mark options={solid}]
table {%
-9 19.5954303741455
-7.5 19.9849643707275
-6 19.6582469940186
-4.5 19.3794460296631
-3 19.1393508911133
-1.5 18.8579235076904
0 18.6162738800049
};
\addplot [semithick, blue, dashed, mark=*, mark size=1, mark options={solid}]
table {%
-9 17.7175712585449
-7.5 15.2317590713501
-6 13.3039426803589
-4.5 11.6705293655396
-3 9.42830181121826
-1.5 6.47690963745117
0 4.33759355545044
};
\addplot [semithick, black, dotted, mark=x, mark size=2, mark options={solid}]
table {%
-9 20.9267463684082
-7.5 18.4329509735107
-6 15.8116016387939
-4.5 13.5395040512085
-3 11.7451887130737
-1.5 10.3244276046753
0 9.23413848876953
};
\draw (-8.5, 41) arc
    [
        start angle=50,
        end angle=310,
        x radius=0.1cm,
        y radius =0.45cm
    ] ;

\draw (-8.5, 20.75) arc
[
    start angle=50,
    end angle=310,
    x radius=0.1cm,
    y radius =0.35cm
] ;
\draw[] (-8.65, 16) -- (-8.85, 14);
\draw[] (-8.65, 35) -- (-8.85, 33);
\node[] (k2) at (-8.5, 13) {\footnotesize$K=2$};
\node[] (k4) at (-8.5,  32) {\footnotesize$K=4$};

\node[circle,draw, inner sep=0pt, minimum size=0.2cm] (c1) at (-6, 30.1344528198242){}; 
\node[circle,draw, inner sep=0pt, minimum size=0.2cm] (c2) at (-3, 30.217586517334){}; 
\draw [densely dotted] (c2.north) -- (-3, 40);
\draw [densely dotted] (c1.north) -- (-6, 40);
\draw[<->] (-3, 40) -- (-6, 40);
\node[] at (-4.5, 42) {\footnotesize \SI{3}{\decibel}};

\end{axis}

\end{tikzpicture}
    \caption{Achievable sum rates averaged over 500 channel realizations of the test set. $M=64$, $K\in\{2, 4\}$, $P_T/\sigma^2_v=20$ \SI{}{\decibel} with varied \gls{ibo}. Comparing the \gls{nn}, \gls{zf} and \gls{zf} plus \gls{dpd}~(\ref{eq:dpd}). The \gls{nn} is retrained at each \gls{ibo} point.} 
    \label{fig:diffbo}
\end{figure}
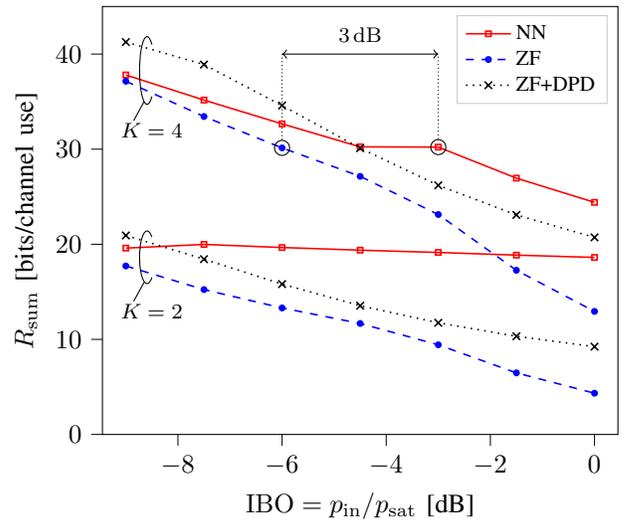

Fig.~\ref{fig:diffbo} depicts the sum rate when $p_{\mathrm{in}}$ is fixed but $p_{\mathrm{sat}}$ is varied, resulting in a varied \gls{ibo}. This shows that, for $K\in\{2,4\}$, the \gls{nn} always outperforms the classical \gls{zf} precoder, when evaluated using the polynomial \gls{pa} model in~(\ref{eq:poly}). For instance, when $K=4$, in order to achieve a sum rate of 30 bits/channel use, the \gls{zf} precoder requires \SI{-6}{\decibel} back-off while the one generated by the \gls{nn} only requires \SI{-3}{\decibel}, implying a significant increase in energy efficiency. Additionally, when $K=2$, the \gls{nn} is able to achieve a nearly constant sum rate over a wide range of \gls{ibo}. This illustrates the ability of the \gls{nn} to suppress nearly all third-order distortion. When comparing the \gls{nn} with \gls{zf} plus a perfect \gls{dpd} in Fig~\ref{fig:diffbo}, it is evident that the \gls{nn} is most beneficial when a lot of distortion is present, i.e., at low back-off. However, we stress the fact that the \gls{nn} does not have to be used as a replacement, but could be used in combination with \gls{dpd}. When (perfect) \gls{dpd} is available, the \gls{pa} characteristic after applying \gls{dpd} can be modeled as a polynomial on which the \gls{nn} can be retrained. As such, the combination of both approaches could produce even better results. 

\section{Conclusion}
In this study, a \gls{ccnn} is trained in a self-supervised manner to learn the mapping between channel matrix and linear precoding matrix in the presence of non-linear \gls{pa} distortion. By learning this mapping, \glspl{pa} can be operated closer to saturation implying a more energy-efficient operating point. Simulation results indicate that the proposed solution outperforms classical precoding schemes such as \gls{zf}. This conclusion holds even when \gls{zf} is combined with perfect \gls{dpd}, given that the system is distortion-limited, i.e., in an energy-efficient regime. These results are especially promising in the multi-user case where no closed-form solutions for the precoder are available. Future perspectives include the use of \glspl{gnn} which enable a lower complexity and allow for the incorporation of additional knowledge. For instance, currently, the system has to be retrained when the \gls{pa} parameters or operating \gls{snr} changes. This could be avoided by incorporating the \gls{snr} and \gls{pa} parameters as inputs to the network. Additionally, the third-order \gls{pa} model is only valid when entering the saturation regime. Future work should adopt more complex \gls{pa} models that capture higher-order and memory effects. Finally, the impact of channel estimation and \gls{pa} parameter estimation errors on the performance of the proposed method has to be evaluated in future studies.

\begin{figure}[t]
    \centering
\begin{tikzpicture}

\definecolor{darkgray176}{RGB}{176,176,176}
\definecolor{darkorange25512714}{RGB}{255,127,14}
\definecolor{forestgreen4416044}{RGB}{44,160,44}
\definecolor{lightgray204}{RGB}{204,204,204}
\definecolor{steelblue31119180}{RGB}{31,119,180}

\begin{axis}[
legend cell align={left},
legend style={nodes={scale=0.8, transform shape},
  fill opacity=0.8,
  draw opacity=1,
  text opacity=1,
  at={(0.03,0.97)},
  anchor=north west,
  draw=lightgray204
},
tick align=outside,
tick pos=left,
x grid style={darkgray176},
xlabel={$P_T/\sigma^2_v$ [\SI{}{\decibel}]},
xmin=-33.25, xmax=38.25,
xtick style={color=black},
y grid style={darkgray176},
ylabel={$R_{\mathrm{sum}}$ [bits/channel use]},
ymin=-1.08993997648757, ymax=33.5,
ytick style={color=black}
]
\addplot [semithick, red]
table {%
-30 0.0306507349014282
-27.1739139556885 0.0584536790847778
-24.3478260040283 0.110965847969055
-21.5217399597168 0.208876609802246
-18.6956520080566 0.387279033660889
-15.8695650100708 0.699925303459167
-13.043478012085 1.21568191051483
-10.2173910140991 1.99868428707123
-7.39130449295044 3.07666635513306
-4.5652174949646 4.42360639572144
-1.73913037776947 5.97530841827393
1.08695650100708 7.66057443618774
3.91304349899292 9.42108821868896
9.56521701812744 13.0064897537231
12.3913040161133 14.7649459838867
15.2173910140991 16.448673248291
18.043478012085 18.0039443969727
20.8695659637451 19.3674831390381
23.6956520080566 20.4821720123291
26.5217399597168 21.320104598999
29.3478260040283 21.8964424133301
32.1739120483398 22.2613887786865
35 22.4773445129395
};
\addlegendentry{NN}
\addplot [thick, black, dotted]
table {%
-30 0.0649948120117188
-27.1739139556885 0.123275756835938
-24.3478260040283 0.231669068336487
-21.5217399597168 0.42824649810791
-18.6956520080566 0.769837617874146
-15.8695650100708 1.3252614736557
-13.043478012085 2.14965391159058
-10.2173910140991 3.24882316589355
-7.39130449295044 4.56448745727539
-1.73913037776947 7.41367721557617
1.08695650100708 8.70656108856201
3.91304349899292 9.77290153503418
6.73913049697876 10.5624446868896
9.56521701812744 11.0885181427002
12.3913040161133 11.4094266891479
15.2173910140991 11.5931282043457
18.043478012085 11.6941061019897
20.8695659637451 11.7483062744141
23.6956520080566 11.7770128250122
29.3478260040283 11.8000183105469
35 11.8063125610352
};
\addlegendentry{ZF + DPD}
\addplot [semithick, blue, dashed]
table {%
-30 0.0542991161346436
-27.1739139556885 0.103126764297485
-24.3478260040283 0.19427478313446
-21.5217399597168 0.360647201538086
-18.6956520080566 0.652782559394836
-15.8695650100708 1.13489806652069
-13.043478012085 1.86291742324829
-10.2173910140991 2.84717988967896
-7.39130449295044 4.02805137634277
-4.5652174949646 5.28825950622559
-1.73913037776947 6.48908758163452
1.08695650100708 7.50961542129517
3.91304349899292 8.28148651123047
6.73913049697876 8.80419158935547
9.56521701812744 9.12690925598145
12.3913040161133 9.31316566467285
15.2173910140991 9.41606712341309
18.043478012085 9.47146034240723
20.8695659637451 9.50084781646729
26.5217399597168 9.52442169189453
35 9.53203105926514
};
\addlegendentry{ZF}

\addplot [semithick, red]
table {%
-30 0.0347353219985962
-27.1739139556885 0.0663948059082031
-24.3478260040283 0.126581430435181
-21.5217399597168 0.240155458450317
-18.6956520080566 0.45156455039978
-15.8695650100708 0.835602521896362
-13.043478012085 1.50507473945618
-10.2173910140991 2.60008645057678
-7.39130449295044 4.24205493927002
-4.5652174949646 6.4672703742981
-1.73913037776947 9.19809341430664
1.08695650100708 12.2810564041138
6.73913049697876 18.8271102905273
9.56521701812744 21.967586517334
12.3913040161133 24.7991790771484
15.2173910140991 27.1638126373291
18.043478012085 28.9623241424561
20.8695659637451 30.2002239227295
23.6956520080566 30.9780941009521
26.5217399597168 31.4333992004395
29.3478260040283 31.6873264312744
32.1739120483398 31.8247966766357
35 31.8979606628418
};
\addplot [thick, black, dotted]
table {%
-30 0.0675326585769653
-27.1739139556885 0.128751635551453
-24.3478260040283 0.244273662567139
-21.5217399597168 0.459299206733704
-18.6956520080566 0.849817991256714
-15.8695650100708 1.53005397319794
-13.043478012085 2.6402862071991
-10.2173910140991 4.29728269577026
-7.39130449295044 6.5241379737854
-4.5652174949646 9.22149276733398
-1.73913037776947 12.2067966461182
1.08695650100708 15.2693691253662
3.91304349899292 18.1956081390381
6.73913049697876 20.7796592712402
9.56521701812744 22.8570709228516
12.3913040161133 24.3608570098877
15.2173910140991 25.3442802429199
18.043478012085 25.9359912872314
20.8695659637451 26.2716846466064
23.6956520080566 26.4552192687988
26.5217399597168 26.5534343719482
29.3478260040283 26.6053695678711
32.1739120483398 26.6326560974121
35 26.6469440460205
};
\addplot [semithick, blue, dashed]
table {%
-30 0.0572081804275513
-27.1739139556885 0.109150290489197
-24.3478260040283 0.207378149032593
-21.5217399597168 0.39093554019928
-18.6956520080566 0.72662615776062
-15.8695650100708 1.31807994842529
-13.043478012085 2.2998411655426
-10.2173910140991 3.79652643203735
-7.39130449295044 5.85139560699463
-4.5652174949646 8.37939739227295
-1.73913037776947 11.1897134780884
1.08695650100708 14.0433807373047
3.91304349899292 16.6968688964844
6.73913049697876 18.9391231536865
9.56521701812744 20.6438579559326
12.3913040161133 21.8088493347168
15.2173910140991 22.5342655181885
18.043478012085 22.9555358886719
20.8695659637451 23.1891860961914
23.6956520080566 23.3152484893799
26.5217399597168 23.382209777832
29.3478260040283 23.4174766540527
35 23.4456386566162
};




\draw (2,16.625) arc
    [
        start angle=50,
        end angle=310,
        x radius=0.1cm,
        y radius =0.425cm
    ] ;

\draw (2,9.6) arc
[
    start angle=50,
    end angle=310,
    x radius=0.1cm,
    y radius =0.25cm
] ;
\draw[] (0.67, 16.55) -- (-1, 17.55);
\draw[] (1.5, 6.95) -- (3.17, 5.95);
\node[] (k2) at (8, 5.95) {\footnotesize$K=2$};
\node[] (k4) at (-6,  17.55) {\footnotesize$K=4$};

\end{axis}
\end{tikzpicture}
    \caption{Achievable rates averaged over 500 channel realizations taken from the test set, for $M=64$, $K\in\{2, 4\}$ and \gls{ibo} = \SI{-3}{\decibel}. Comparing the \gls{nn} precoder against \gls{zf} and \gls{zf} plus a perfect \gls{dpd}~(\ref{eq:dpd}).}
    \label{fig:multiuser}
\end{figure}
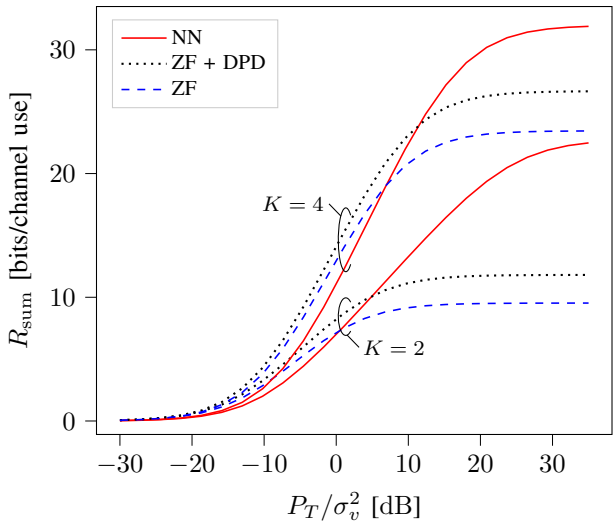


\linespread{0.95}

\bibliographystyle{IEEEtran}
\bibliography{IEEEabrv,mybib}

\begin{thebibliography}{10}
\providecommand{\url}[1]{#1}
\csname url@samestyle\endcsname
\providecommand{\newblock}{\relax}
\providecommand{\bibinfo}[2]{#2}
\providecommand{\BIBentrySTDinterwordspacing}{\spaceskip=0pt\relax}
\providecommand{\BIBentryALTinterwordstretchfactor}{4}
\providecommand{\BIBentryALTinterwordspacing}{\spaceskip=\fontdimen2\font plus
\BIBentryALTinterwordstretchfactor\fontdimen3\font minus
  \fontdimen4\font\relax}
\providecommand{\BIBforeignlanguage}[2]{{%
\expandafter\ifx\csname l@#1\endcsname\relax
\typeout{** WARNING: IEEEtran.bst: No hyphenation pattern has been}%
\typeout{** loaded for the language `#1'. Using the pattern for}%
\typeout{** the default language instead.}%
\else
\language=\csname l@#1\endcsname
\fi
#2}}
\providecommand{\BIBdecl}{\relax}
\BIBdecl

\bibitem{trends2040}
L.~Belkhir and A.~Elmeligi, ``\BIBforeignlanguage{en}{Assessing {ICT} global
  emissions footprint: {Trends} to 2040 \& recommendations},''
  \emph{\BIBforeignlanguage{en}{Journal of Cleaner Production}}, vol. 177, pp.
  448--463, Mar. 2018.

\bibitem{greendeal}
{European Commission}, ``{The European Green Deal},'' \emph{COM (2019)},
  November 2019.

\bibitem{sdgs}
{United Nations}, ``{The 2030 Agenda and the Sustainable Development Goals: An
  opportunity for Latin America and the Caribbean},'' {(LC/G.2681-P/Rev.3),
  Santiago, 2018}.

\bibitem{energy}
G.~Auer, V.~Giannini, C.~Desset, I.~Godor, P.~Skillermark, M.~Olsson, M.~A.
  Imran, D.~Sabella, M.~J. Gonzalez, O.~Blume, and A.~Fehske,
  ``\BIBforeignlanguage{eng}{{How much energy is needed to run a wireless
  network?}}'' \emph{\BIBforeignlanguage{eng}{IEEE wireless communications}},
  vol.~18, no.~5, pp. 40--49, 2011.

\bibitem{pa_percentage}
H.~Bogucka and A.~Conti, ``{Degrees of freedom for energy savings in practical
  adaptive wireless systems},'' \emph{IEEE Communications Magazine}, vol.~49,
  no.~6, pp. 38--45, 2011.

\bibitem{pa_for_wireless}
S.~Cripps, \emph{\BIBforeignlanguage{eng}{{RF Power Amplifiers for Wireless
  Communications}}}, ser. Artech House Microwave Library.\hskip 1em plus 0.5em
  minus 0.4em\relax Norwood: Artech House, 2006.

\bibitem{z3ro}
F.~Rottenberg, G.~Callebaut, and L.~Van~der Perre, ``{Z3RO Precoder Canceling
  Nonlinear Power Amplifier Distortion in Large Array Systems},'' in \emph{ICC
  2022 - IEEE International Conference on Communications}, 2022, pp. 432--437.

\bibitem{zerofamily}
------, ``{The Z3RO Family of Precoders Cancelling Nonlinear Power
  Amplification Distortion in Large Array Systems},'' \emph{IEEE Transactions
  on Wireless Communications}, pp. 1--1, 2022.

\bibitem{z3ro_val}
T.~Feys, G.~Callebaut, L.~Van~der Perre, and F.~Rottenberg,
  ``{Measurement-Based Validation of Z3RO Precoder to Prevent Nonlinear
  Amplifier Distortion in Massive MIMO Systems},'' in \emph{2022 IEEE 95th
  Vehicular Technology Conference: (VTC2022-Spring)}, 2022, pp. 1--5.

\bibitem{distortion-aware}
\BIBentryALTinterwordspacing
S.~R. Aghdam, S.~Jacobsson, U.~Gustavsson, G.~Durisi, C.~Studer, and
  T.~Eriksson, ``{Distortion-Aware Linear Precoding for Massive MIMO Downlink
  Systems with Nonlinear Power Amplifiers},'' 2020. [Online]. Available:
  \url{https://arxiv.org/abs/2012.13337}
\BIBentrySTDinterwordspacing

\bibitem{pas}
T.~Schenk, \emph{\BIBforeignlanguage{eng}{{RF Imperfections in High-rate
  Wireless Systems: Impact and Digital Compensation}}}, 1st~ed.\hskip 1em plus
  0.5em minus 0.4em\relax Dordrecht: Springer Netherlands, 2008.

\bibitem{demir2020bussgang}
O.~T. Demir and E.~Bjornson, ``{The Bussgang Decomposition of Nonlinear
  Systems: Basic Theory and MIMO Extensions [Lecture Notes]},'' \emph{IEEE
  Signal Processing Magazine}, vol.~38, no.~1, pp. 131--136, 2021.

\bibitem{inductivebias}
\BIBentryALTinterwordspacing
P.~W. Battaglia \emph{et~al.}, ``{Relational inductive biases, deep learning,
  and graph networks},'' 2018. [Online]. Available:
  \url{https://arxiv.org/abs/1806.01261}
\BIBentrySTDinterwordspacing

\bibitem{cnn_gnn}
B.~Zhao, J.~Guo, and C.~Yang, ``{Learning Precoding Policy: CNN or GNN?}'' in
  \emph{2022 IEEE Wireless Communications and Networking Conference (WCNC)},
  2022, pp. 1027--1032.

\bibitem{resnet}
\BIBentryALTinterwordspacing
G.~Philipp, D.~Song, and J.~G. Carbonell, ``Gradients explode - deep networks
  are shallow - resnet explained,'' 2018. [Online]. Available:
  \url{https://openreview.net/forum?id=HkpYwMZRb}
\BIBentrySTDinterwordspacing

\bibitem{receptivefield}
A.~Araujo, W.~Norris, and J.~Sim, ``Computing receptive fields of convolutional
  neural networks,'' \emph{Distill}, 2019,
  https://distill.pub/2019/computing-receptive-fields.

\bibitem{adam}
\BIBentryALTinterwordspacing
D.~P. Kingma and J.~Ba, ``Adam: A method for stochastic optimization,'' 2014.
  [Online]. Available: \url{https://arxiv.org/abs/1412.6980}
\BIBentrySTDinterwordspacing

\bibitem{modified_rapp}
C.-S. Choi \emph{et~al.}, ``\BIBforeignlanguage{en}{{RF impairment models for
  60GHz-band SYS/PHY simulation}},'' \emph{\BIBforeignlanguage{en}{Project:
  {IEEE} {P802}.15 {Working} {Group} for {Wireless} {Personal} {Area}
  {Networks} ({WPANs})}}, p.~17, 2006.

\bibitem{3gpp}
{Nokia}, ``{Realistic power amplifier model for the New Radio evaluation},''
  \emph{3GPP TSG-RAN WG4 Meeting 79, R4-163314}, May 2016.

\bibitem{distortion_beamformed2}
C.~Mollen, U.~Gustavsson, T.~Eriksson, and E.~G. Larsson,
  ``\BIBforeignlanguage{eng}{{Spatial Characteristics of Distortion Radiated
  From Antenna Arrays With Transceiver Nonlinearities}},''
  \emph{\BIBforeignlanguage{eng}{IEEE transactions on wireless
  communications}}, vol.~17, no.~10, pp. 6663--6679, 2018.

\end{thebibliography}

\end{document}